# LIME: Live Intrinsic Material Estimation

Abhimitra Meka [1,2]    Maxim Maximov [1,2]    Michael Zollhöfer [1,2,3]    Avishek Chatterjee [1,2]
Hans-Peter Seidel [1,2]    Christian Richardt [4]    Christian Theobalt [1,2]

[1] MPI Informatics    [2] Saarland Informatics Campus    [3] Stanford University    [4] University of Bath

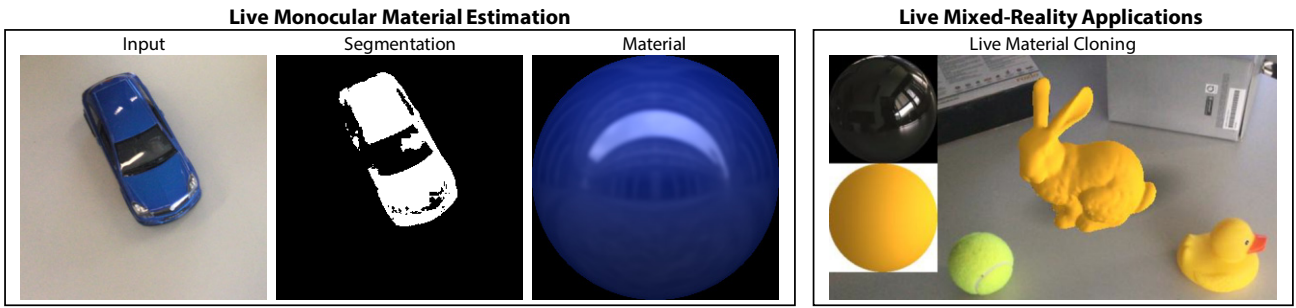

Figure 1. Our approach enables the real-time estimation of the material of general objects (left) from just a single monocular color image. This enables exciting live mixed-reality applications (right), such as for example cloning a real-world material onto a virtual object.

## Abstract

*We present the first end-to-end approach for real-time material estimation for general object shapes with uniform material that only requires a single color image as input. In addition to Lambertian surface properties, our approach fully automatically computes the specular albedo, material shininess, and a foreground segmentation. We tackle this challenging and ill-posed inverse rendering problem using recent advances in image-to-image translation techniques based on deep convolutional encoder–decoder architectures. The underlying core representations of our approach are specular shading, diffuse shading and mirror images, which allow to learn the effective and accurate separation of diffuse and specular albedo. In addition, we propose a novel highly efficient perceptual rendering loss that mimics real-world image formation and obtains intermediate results even during run time. The estimation of material parameters at real-time frame rates enables exciting mixed-reality applications, such as seamless illumination-consistent integration of virtual objects into real-world scenes, and virtual material cloning. We demonstrate our approach in a live setup, compare it to the state of the art, and demonstrate its effectiveness through quantitative and qualitative evaluation.*

## 1. Introduction

The estimation of material properties from a single monocular color image is a high-dimensional and underconstrained problem. The blind deconvolution nature of the problem has attracted usage of complex setups and, more recently, various natural and handcrafted priors, but has yet remained outside the scope of real-time implementation due to the resulting dense optimization problem. Previous real-time approaches have thus predominantly focused on estimating diffuse materials [31, 33]. In this work, we tackle a much harder inverse problem by additionally estimating specular material properties, such as specular color and material shininess, as well as segmentation masks for general objects of uniform material from a single color image or video in real time.

Recent advances in deep learning enable the automatic learning of underlying natural subspace constraints directly from large training data, while also reducing the need to solve the expensive dense non-linear optimization problem directly. Some recent work has successfully demonstrated the capability of convolutional neural networks to solve the inverse rendering problem of separating material from illumination, particularly in the context of single material objects. Current approaches estimate material from one [12, 28] or more images [21, 44]. Georgoulis et al. [12] learn BRDF parameters and outdoor environment maps from single images of specular objects from a specific class (cars, chairs and couches only). Kim et al. [21] estimate BRDF parameters from multiple RGB input images in 90 ms. Shi et al. [41] perform intrinsic image decomposition of a single object image into diffuse and specular layers but do not solve the denser and more complex material estimation problem.

Most of these methods take a direct approach to parameter regression without any additional supervision, due to which the network may not necessarily learn to perform the physical deconvolution operation that is intrinsic to inverse rendering, and hence runs the risk of simply overfitting to the training data. The exception is the approach of Liu et al. [28] that took a first important step in this direction using an expert-designed rendering layer. However, such a rendering layer requires shape estimation in the form of surface normals, which are challenging to regress for general objects. This limits their method to objects of particular shape classes (cars, chairs and couches), and also requires manual segmentation of the object in the image.

In contrast, we present the first real-time material estimation method that works for objects of any general shape, and without manual segmentation, thus making our approach applicable to live application scenarios. Our approach draws inspiration from the rendering process. We decompose the input image into intrinsic image layers and provide fine-grained intermediate supervision by following the rendering process closely. We decouple the task of material estimation from the shape estimation problem by introducing a novel image-space supervision strategy on the intrinsic layers using a highly efficient perceptual loss that makes direct use of the regressed layers. We finally regress each material parameter from the relevant intrinsic image layers, self-supervised by the perceptual rendering loss. This mechanism results in demonstrably more accurate material estimation.

In addition to these core innovations, we distinguish ourselves from previous work in the following ways:

1. We fully automatically perform object segmentation in the image, enabling our method to be applied to single images and videos, also in live scenarios.
2. We train our network for the challenging indoor setting, and successfully handle complex high-frequency lighting as opposed to the natural outdoor illumination used by other methods [12, 21, 28], since most mixed-reality applications are used indoors.
3. If shape information is available, e.g. from a depth sensor, our method also extracts separate low- and high-frequency lighting information, which is crucial for vivid AR applications.

## 2. Related Work

The appearance of an object in an image depends on its surface geometry, material and illumination. Estimation of these components is a fundamental problem in computer vision, and joint estimation the ultimate quest of inverse rendering [36, 46]. Geometry reconstruction has seen major advances since the release of commodity depth sensors [e.g. 5, 16, 18, 35, 47]. However, estimation of material and illumination remains relatively more challenging. Approaches for estimating material and illumination need to make strong assumptions, such as the availability of a depth sensor [13, 39], lighting conditions corresponding to photometric stereo [15], a rotating object under static illumination [45], multiple images of the same object under varying illumination [42], having an object of a given class [12], or requiring user input [34].

**Material Estimation** There are broadly two classes of material estimation approaches: (1) approaches that assume known geometry, and (2) approaches for specific object classes of unknown geometry. Methods that require the surface geometry of objects to be known can, in principle, work on any type of surface geometry. Dong et al. [9] estimate spatially-varying reflectance from the video of a rotating object of known geometry. Wu and Zhou [43] perform on-the-fly appearance estimation by exploiting the infrared emitter–receiver system of a Kinect as an active reflectometer. Knecht et al. [23] also propose a method for material estimation at interactive frame rates using a Kinect sensor. Li et al. [26] learn surface appearance of planar surfaces from single images using self-augmented CNNs. There are also several recent off-line methods [21, 38, 44] that capture a set of RGB images along with aligned depth maps to estimate an appearance model for the surface geometry. Recent methods by Rematas et al. [37], Georgoulis et al. [12] and Liu et al. [28] do not assume known geometry, but instead rely on implicit priors about the object shape, and therefore only work on the specific classes of objects – such as cars or chairs – for which the methods are trained.

In contrast to these methods, our approach neither requires known surface geometry nor is it restricted to specific object classes. To the best of our knowledge, the only other RGB-only method that works on arbitrary objects is by Lombardi and Nishino [29]. However, it is an offline method. We believe our real-time method can significantly enhance a wide variety of applications like material editing [3, 7, 8, 20], object relighting [27], cloning and insertion.

**Illumination Estimation** Assuming a diffuse reflectance, Marschner and Greenberg [32] estimate environment maps from captured RGB images and scanned geometry. Given a single input image, methods exist for estimating natural outdoor illumination [14, 25], indoor illumination [10] or the location of multiple light sources [30]. Georgoulis et al. [11] estimate an environment map from the photo of a multicolored specular object of known shape. Mandl et al. [31] similarly learn the lighting from a single image of a known object. Lalonde and Matthews [24] perform illumination estimation from an image collection used for structure-from-motion reconstruction. However, note that the main contribution of this paper lies in material estimation, and not illumination estimation. Nevertheless, given geometry, we show in Section 7 how our approach can be extended to additionally estimate illumination.

## 3. Overview

Our approach is the first end-to-end approach for real-time estimation of an object's material and segmentation mask from just a single color image. We start by introducing our image formation model in Section 4. In Section 5, we discuss how we tackle the underlying inverse rendering problem using en-

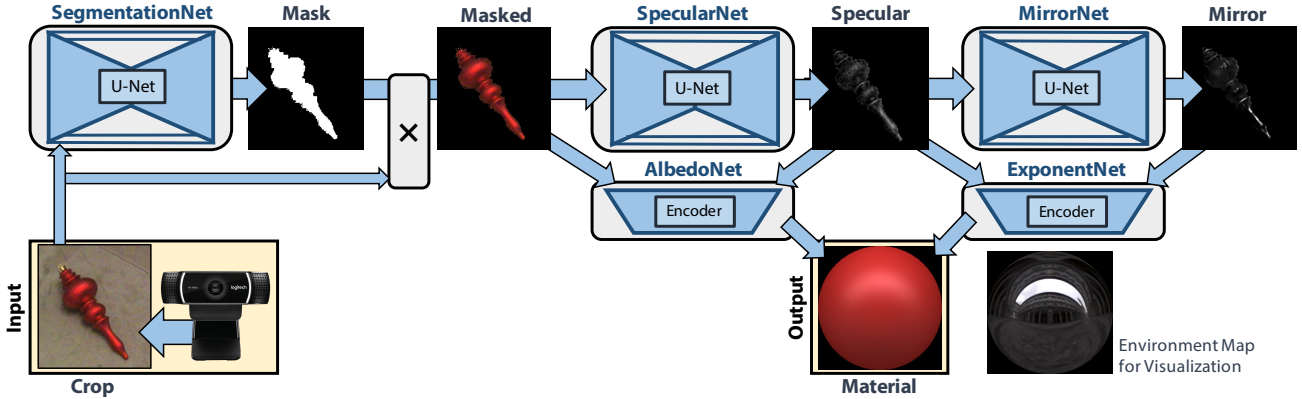

Figure 2. Our approach enables real-time estimation of material parameters from a single monocular color image (bottom left). The proposed end-to-end learning approach decomposes the complex inverse rendering problem into sub-parts that are inspired by the physical real-world image formation process, leading to five specifically tailored subnetworks. Our complete network is trained in an end-to-end fashion. Environment map used for material visualization (bottom right) courtesy of Debevec [6].

coder–decoder architectures [17, 40]. Our network is inspired by the image formation process, and thus the quantities involved in the rendering equation [19]. In particular, we use the Blinn–Phong reflection model [2] that allows for the observed image radiance to be decomposed linearly into a diffuse and a specular layer, which can be further decomposed to provide the corresponding shading layers and albedos. We estimate such a shading decomposition and albedos, and recombine them to obtain a rendering loss between the reconstructed image and the input image, to supervise the network training.

Figure 2 outlines our end-to-end approach and its five specifically tailored subnetworks: *SegmentationNet* estimates a binary object mask. *SpecularNet* decomposes the masked input to obtain the specular shading image, which quantifies the normalized specular reflections from the object. *MirrorNet* converts the specular shading into a 'mirror image', a novel representation that quantifies the incoming high-frequency illumination onto the object's surface. We call it 'mirror image' because it captures how the object would look if it were a perfectly reflective mirror-like surface, see Figure 3. Such a representation maps the environmental illumination to the image space, allowing for an easier estimation task for the high-frequency lighting. *AlbedoNet* uses the masked input and the estimated specular shading to regress the linear diffuse and specular albedo values, and *ExponentNet* uses the specular shading and the mirror image to regress the non-linear specular exponent.

Structuring our architecture in this manner gives the opportunity for intermediate supervision of the involved physical quantities, which results in higher-quality results than competing approaches, as we show in our results section. In addition, our architecture enables the computation of a perceptual rendering loss, which leads to higher-quality results. For higher temporal stability, when the method is applied to video, the reconstructed material parameters are temporally fused. We show and evaluate our results, and compare to state-of-the-art techniques in Section 6. Finally, in Section 7, we demonstrate mixed-reality applications that benefit from our real-time inverse rendering approach, such as seamless placement of virtual objects, with real-time captured materials, in real-world scenes.

## 4. Image Formation Model

Obtaining diffuse and specular material parameters requires the inversion of the complex real-world image formation process. In this section, we thus explain the forward process of image formation and all employed scene assumptions.

### 4.1. Appearance and Illumination Model

The appearance of an object in an image depends on its bidirectional reflectance distribution function (BRDF) and the light transport in the scene. We model light transport based on the trichromatic approximation of the rendering equation [19]:

$$\mathbf{L}_o(\mathbf{x},\boldsymbol{\omega}_o) = \mathbf{L}_e(\mathbf{x},\boldsymbol{\omega}_o) + \mathbf{L}_r(\mathbf{x},\boldsymbol{\omega}_o)$$
$$= \mathbf{L}_e(\mathbf{x},\boldsymbol{\omega}_o) + \int_\Omega \mathbf{f}(\mathbf{x},\boldsymbol{\omega}_i,\boldsymbol{\omega}_o)\mathbf{L}_i(\mathbf{x},\boldsymbol{\omega}_i)(\boldsymbol{\omega}_i\cdot\mathbf{n})\,d\boldsymbol{\omega}_i. \quad (1)$$

The rendering equation expresses the radiance $\mathbf{L}_o \in \mathbb{R}^3$ leaving a surface point $\mathbf{x} \in \mathbb{R}^3$ (with normal $\mathbf{n} \in \mathbb{S}^2$) in direction $\boldsymbol{\omega}_o \in \mathbb{S}^2$ as the sum of emitted $\mathbf{L}_e \in \mathbb{R}^3$ and reflected radiance $\mathbf{L}_r \in \mathbb{R}^3$. The reflected radiance $\mathbf{L}_r$ is a function of the illumination $\mathbf{L}_i \in \mathbb{R}^3$ over the hemisphere $\Omega$ of incoming directions $\boldsymbol{\omega}_i \in \mathbb{S}^2$ and the material's BRDF $\mathbf{f} : \mathbb{R}^3 \times \mathbb{S}^2 \times \mathbb{S}^2 \to \mathbb{R}$ at point $\mathbf{x}$.

To make real-time inverse rendering tractable, we make a few simplifying assumptions, which are widely used, even in off-line state-of-the-art inverse rendering techniques [12, 28]. First, we assume that the object is not emitting light, i.e., it is not a light source, and we only model direct illumination. We model global changes in scene brightness based on an ambient illumination term $\mathbf{L}_a \in \mathbb{R}^3$. We further assume distant lighting and the absence of self-shadowing, which decouples the incident illumination from the object's spatial embedding.

Given these assumptions, the rendering equation simplifies to

$$\mathbf{L}(\mathbf{x},\boldsymbol{\omega}_o)=\mathbf{L}_a+\int_\Omega\underbrace{\mathbf{f}(\mathbf{x},\boldsymbol{\omega}_i,\boldsymbol{\omega}_o)(\boldsymbol{\omega}_i\cdot\mathbf{n})}_{\mathbf{BP}(\mathbf{x},\mathbf{n},\boldsymbol{\omega}_i,\boldsymbol{\omega}_o)}\mathbf{E}(\boldsymbol{\omega}_i)d\boldsymbol{\omega}_i. \quad (2)$$

We represent distant illumination using an environment map $\mathbf{E}(\boldsymbol{\omega}_i)$. Diffuse and specular object appearance is parameterized using the Blinn–Phong reflection model [2]:

$$\mathbf{BP}(\mathbf{x},\mathbf{n},\boldsymbol{\omega}_i,\boldsymbol{\omega}_o)=\underbrace{\mathbf{m}_d(\boldsymbol{\omega}_i\cdot\mathbf{n})}_{\text{diffuse}}+\underbrace{\mathbf{m}_s(\mathbf{h}\cdot\mathbf{n})^s}_{\text{specular}}. \quad (3)$$

Here, $\mathbf{m}_d \in \mathbb{R}^3$ is the diffuse, and $\mathbf{m}_s \in \mathbb{R}^3$ the specular material color (albedo). Note that we assume a white specularity, i.e., $\mathbf{m}_s = \alpha \cdot \mathbf{1}_3$, with $\mathbf{1}_3$ being a 3-vector of ones. The halfway vector $\mathbf{h} = \frac{\boldsymbol{\omega}_i+\boldsymbol{\omega}_o}{\|\boldsymbol{\omega}_i+\boldsymbol{\omega}_o\|}$ depends on the light direction $\boldsymbol{\omega}_i$ and the viewing direction $\boldsymbol{\omega}_o$. The scalar exponent $s \in \mathbb{R}$ determines the size of the specular lobe, and thus the shininess of the material.

## 5. Deep Material Learning

The goal of our approach is the real-time estimation of diffuse and specular object material from a single color image. This high-dimensional and non-linear inverse rendering problem is ill-posed, since each single color measurement is the integral over the hemisphere of the product between the BRDF and the incident illumination modulated by the unknown scene geometry (see Equation 1).

We propose a novel discriminative approach to tackle this challenging problem using deep convolutional encoder–decoder architectures. Our network is inspired by the quantities involved in the physical image formation process. Structuring the CNN architecture in this way gives the opportunity for intermediate supervision of the physical quantities and leads to higher-quality results than other competing approaches, as shown in Section 6. In addition, this architecture facilitates the computation of a perceptual rendering loss, which further improves regression results. In the following, we describe our synthetic ground-truth training corpus, our physically motivated inverse rendering network, a novel perceptual per-pixel rendering loss, and show how our entire network can be trained end-to-end.

### 5.1. Synthetic Ground-Truth Training Corpus

Since the annotation of real-world images with ground-truth BRDF parameters is practically infeasible, we train our deep networks on fully synthetically generated imagery with readily available ground truth. Our training corpus

$$\mathcal{T}=\{\mathbf{I}_i,\mathbf{B}_i,\mathbf{D}_i,\mathbf{S}_i,\mathbf{M}_i,\mathbf{BP}_i\}_{i=1}^{N}$$

consists of $N=100{,}000$ realistically rendered images $\mathbf{I}_i$, their corresponding binary segmentation masks $\mathbf{B}_i$, diffuse shading images $\mathbf{D}_i$, specular shading images $\mathbf{S}_i$, mirror images $\mathbf{M}_i$, and the ground-truth Blinn–Phong parameters $\mathbf{BP}_i$. See Figure 3 for examples from our corpus, and our supplemental document for more examples.

Each of the $N$ training frames shows a single randomly sampled object from a set of 55 synthetic 3D models[1,2] (50 models for training and 5 for testing). We render the object with random pose, orientation, size and Blinn–Phong parameters $\mathbf{BP}_i$ using perspective projection to obtain our training corpus $\mathcal{T}$. The albedo parameters are sampled uniformly in the YUV color space and then converted to RGB.

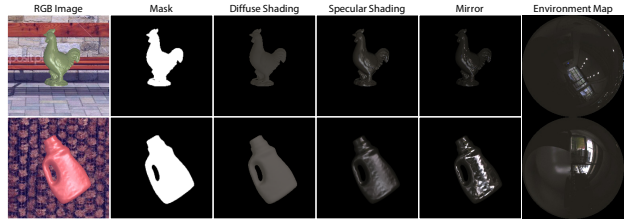

Figure 3. Two examples from our synthetic ground-truth training corpus (from left to right): color image $\mathbf{I}$, segmentation mask $\mathbf{B}$, diffuse $\mathbf{D}$ and specular $\mathbf{S}$ shading image, mirror image $\mathbf{M}$, and environment map $\mathbf{E}$. Contrast of $\mathbf{E}$ increased for better visualization.

The object is lit with a spherical environment map $\mathbf{E}_i$, which we randomly sample from a set of 45 indoor maps that we captured with an LG 360 Cam with manual exposure control, see Figure 3 (right). The environment maps were captured in varied indoor settings, in rooms of different sizes and different lighting arrangement, such as homes, offices, classrooms and auditoriums. For data augmentation, we randomly rotate environment maps while ensuring there is a strong light source in the frontal hemisphere. This ensures that highlights will be visible if the object is specular.

We render objects under different perspective views and obtain crops around the objects at different resolutions with varying amounts of translation and scaling. We add a background based on random textures to the rendered object image to provide sufficient variety for the segmentation network to learn foreground segmentation. Our training corpus will be made publicly available.[3]

### 5.2. Physically Motivated Network Architecture

The proposed network architecture is inspired by the physical image formation process (Section 4), and thus the quantities involved in the rendering equation, as illustrated in Figure 2. We partition the task of material estimation into five CNNs tailored to perform specific sub-tasks. We start with the estimation of a binary segmentation mask (*SegmentationNet*) to identify the pixels that belong to the dominant object in the scene. Afterwards, we decompose the masked input image to obtain the specular shading image (*SpecularNet*). The mirror estimation subnetwork (*MirrorNet*) converts the specular shading image into a mirror image by removing the specular roughness. Finally, the albedo estimation network (*AlbedoNet*) uses the masked input image and the specular shading image to estimate the diffuse and specular albedo parameters. The exponent estimation network (*ExponentNet*) combines the specular shading image and the mirror image to produce the specular exponent, which ranges from diffuse to shiny.

---
[1] http://rll.berkeley.edu/bigbird/
[2] https://www.shapenet.org/about
[3] http://gvv.mpi-inf.mpg.de/projects/LIME/

Our proposed architecture provides the opportunity for intermediate supervision using the known ground-truth quantities from our corpus, which leads to higher-quality regression results than direct estimation with a single CNN. Our approach also enables the implementation of a novel perceptual rendering loss, which we discuss in Section 5.3. While our network is based on five sub-tasks, we do train it end-to-end, which typically results in better performance than using individually designed components. As shown in the results section, our core representation, which is based on specular and mirror images, is better suited for the image-to-image translation task than the direct regression of reflectance maps in previous work [12]. The main reason is that the corresponding image-to-image translation task is easier, in the sense that the CNN has only to learn a per-pixel color function, instead of a color transform in combination with a spatial reordering of the pixel, as is the case for reflectance and environment maps. This is because the pixel locations in reflectance and environment maps inherently depend on the underlying unknown scene geometry of the real-world object. The estimated mirror image, in combination with the specular image, enables the regression of material shininess with higher accuracy, since it provides a baseline for exponent estimation.

The input to our novel inverse rendering network are 256×256 images that contain the full object at the center. The architectures of *SegmentationNet*, *SpecularNet* and *MirrorNet* follow U-Net [40]. The skip connections allow for high-quality image-to-image translation. *AlbedoNet* is an encoder with 5 convolution layers, each followed by ReLU and max-pooling, and 3 fully-connected layers. *ExponentNet* is a classification network that uses a one-hot encoding of the eight possible classes of object shininess. We use binned shininess classes to represent just-noticeably different shininess levels, as regression of scalar (log) shininess exhibited bias towards shiny materials (see Section 6 for discussion). For full details of the used subnetworks, please refer to the supplemental document. During training, we apply an $\ell_2$-loss with respect to the ground truth on all intermediate physical quantities and the output material parameters, except for *SpecularNet* and *MirrorNet*, for which we use an $\ell_1$-loss to achieve high-frequency results, and a cross-entropy loss for classification of shininess using *ExponentNet*. In addition, to further improve decomposition results, we apply a novel perceptual rendering loss, which we describe next.

### 5.3. Perceptual Rendering Loss

Since we train our approach on a synthetic training corpus (see Figure 3), we have ground-truth annotations for all involved physical quantities readily available, also for the ground-truth Blinn–Phong parameters $\mathbf{BP}_i$. One straightforward way of defining a loss function for the material parameters is directly in parameter space, e.g., using an $\ell_2$-loss. We show that this alone is not sufficient, since it is unclear how to optimally distribute the error between the different parameter dimensions, such that the parameter error matches the perceptual per-pixel distance between the ground truth and the corresponding re-rendering of the object. Another substantial drawback of independent parameter space loss functions is that the regression results are not necessarily consistent, i.e., the re-rendering of the object based on the regressed parameters perceptually may not match the input image, since errors in the independent components accumulate. To alleviate these two substantial problems, which are caused by independent parameter loss functions, we propose an additional perceptual rendering loss that leads to results of higher quality. We show the effectiveness of this additional constraint in Section 6.

Our novel perceptual loss is based on rewriting the rendering equation (Equation 1) in terms of the diffuse shading $\mathbf{D}$ and the specular shading $\mathbf{S}$:

$$\mathbf{L}(\mathbf{x},\boldsymbol{\omega}_o) = \mathbf{m}_a + \mathbf{m}_d \underbrace{\int_\Omega (\boldsymbol{\omega}_i \cdot \mathbf{n}) \mathbf{E}(\boldsymbol{\omega}_i)\, d\boldsymbol{\omega}_i}_{\mathbf{D}}$$
$$+ \mathbf{m}_s \underbrace{\int_\Omega (\mathbf{h} \cdot \mathbf{n})^s \mathbf{E}(\boldsymbol{\omega}_i)\, d\boldsymbol{\omega}_i}_{\mathbf{S}}$$
$$= \mathbf{m}_a + \mathbf{m}_d \mathbf{D} + \mathbf{m}_s \mathbf{S}. \qquad (4)$$

For high efficiency during training, we pre-compute the diffuse and specular shading integrals per pixel in our ground-truth training corpus (see Figure 3), and store them in the form of diffuse shading and specular shading maps $\mathbf{D}$ and $\mathbf{S}$, respectively.

Our perceptual rendering loss $\mathbf{R}$ directly measures the distance between the rendered prediction and the input image $\mathbf{I}$:

$$\mathbf{R}(\hat{\mathbf{m}}_d, \hat{\mathbf{m}}_s, \hat{\mathbf{S}}) = \left\| \mathbf{B} \cdot \left[ \mathbf{I} - \underbrace{(\mathbf{A} + \hat{\mathbf{m}}_d \mathbf{D} + \hat{\mathbf{m}}_s \hat{\mathbf{S}})}_{\text{rendered prediction}} \right] \right\|. \qquad (5)$$

Here, $\mathbf{B}$ is the binary foreground mask, and the rendered prediction is based on the ambient color $\mathbf{A} = m_a \mathbf{1}_3$, the predicted diffuse albedo $\hat{\mathbf{m}}_d$, the ground-truth diffuse shading $\mathbf{D}$, the predicted specular albedo $\hat{\mathbf{m}}_s$ and specular shading $\hat{\mathbf{S}}$. We directly predict the specular shading $\hat{\mathbf{S}}$ instead of the shininess $s$ to alleviate the costly integration step over the environment map. Since we pre-computed all physical quantities in our training corpus, the rendering step is a simple per-pixel operation that is highly efficient and can be implemented using off-the-shelf operations such as per-pixel addition and multiplication, which are already provided by deep-learning libraries, without the need for a hand-crafted differentiable rendering engine [28].

### 5.4. End-to-End Training

We train all our networks using TensorFlow [1] with Keras [4]. For fast convergence, we train our novel inverse rendering network in two stages: First, we train all subnetworks separately based on the synthetic ground-truth training corpus. Then we train *SpecularNet* and *MirrorNet* together. Afterwards, we add *ExponentNet*, and finally *AlbedoNet* to the end-to-end training. The gradients for back-propagation are obtained using Adam [22] with default parameters. We first train for 100,000 iterations with a batch size of 32, and then fine-tune end-to-end for 45,000 iterations, with a base learning rate of 0.0001 and $\delta = 10^{-6}$.

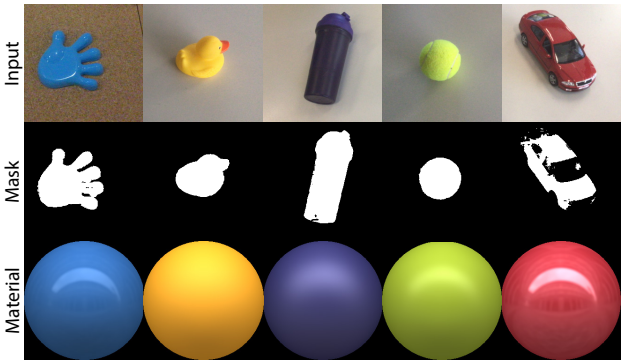

Figure 4. Real-world material estimation results based on a single color image. Our approach produces high-quality results for a large variety of objects and materials, without manual interaction.

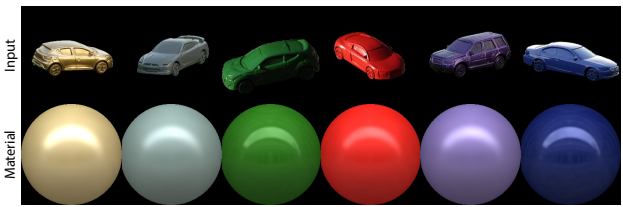

Figure 5. Material estimates on the dataset of Rematas et al. [37].

### 5.5. Temporal Fusion

Our single-shot inverse rendering approach estimates plausible material parameters from a single image. However, when applied to video streams independently per video frame, our estimation may have some temporal instability due to changing lighting conditions, camera parameters or imaging noise. To improve the accuracy and temporal stability of our approach, we therefore propose to temporally fuse all our estimated parameters. This leads to results of higher quality and higher temporal stability. We use a sliding window median filter with a window size of 5 frames. This helps to filter out occasional outliers. From the median-filtered output, we then perform decaying exponential averaging:

$$\mathbf{P}^t = \alpha \widehat{\mathbf{P}} + (1-\alpha) \mathbf{P}^{t-1}. \qquad (6)$$

Here, $\widehat{\mathbf{P}}$ is the current parameter estimate, $\mathbf{P}^t$ is the final estimate for the current time step $t$, and $\mathbf{P}^{t-1}$ is the fused result of the previous frame. We use a decaying blending factor $\alpha = (1/t)$ for all our experiments. This temporal filtering and fusion approach is particularly useful for our environment map estimation strategy (see Section 7), since it helps in integrating novel lighting directions sampled by the object as the camera pans during the video capture.

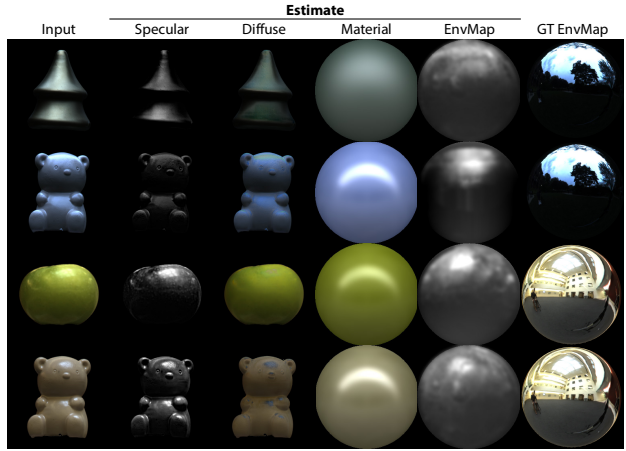

Figure 6. Our approach estimates the specular decomposition layer from a single color image. The diffuse layer can be obtained by subtracting it from the input. With the available ground-truth normals, we also reconstruct the environment map using our technique described in Section 7. Images from Lombardi and Nishino [29].

## 6. Results

We now show qualitative and quantitative results of our real-time single-shot material estimation approach, compare to state-of-the-art approaches, and finally evaluate our design decisions to show the benefits of our novel perceptual loss and physically-motivated network architecture.

**Qualitative Results** Figure 4 shows real-time material estimation results for a wide range of different materials and general objects. As can be seen, our approach estimates material parameters at high quality for many challenging real-world objects that have uniform material, without the need for manual interaction.

We also applied our approach to the photos of painted toy cars by Rematas et al. [37], shown in Figure 5, and obtain high-quality material estimates. In addition, our approach can estimate the specular shading layer from a single color image, which enables us to compute the diffuse shading layer by subtraction, as shown in Figure 6. Note that our approach works for general objects, and does not require manual segmentation. In contrast, previous techniques either work only for a specific object class or require known segmentation [12, 28, 29].

**Run-time Performance** On an Nvidia Titan Xp, a forward pass of our complete inverse rendering network takes 13.72 ms, which enables various live applications discussed in Section 7. Individual run times are: *SegmentationNet* (2.83 ms), *SpecularNet* (3.30 ms), *MirrorNet* (2.99 ms), *AlbedoNet* (2.68 ms) and *ExponentNet* (1.92 ms).

### 6.1. Quantitative Evaluation and Ablation Study

We quantitatively analyze our method's performance to validate our design choices. We compare average estimation errors for groups of material parameters on an unseen test set of 4,990 synthetic images in Table 1. We compare our full approach to three alternative versions by modifying one aspect of the network in each instance, plus one alternative:

1. Our network as-is, but without the novel perceptual loss (Section 5.3). Exclusion of the perceptual loss leads to reduced accuracy in the albedo estimates.

2. Our network without the *MirrorNet*, so that the *ExponentNet* only depends on the output of *SpecularNet*. The exclusion of *MirrorNet* leads to reduced exponent classification accuracy, thus proving the efficacy of our mirror-representation-based design on the challenging task of estimating the non-linear material shininess.

3. Our network with the *ExponentNet* modified to regress shininess directly instead of as a classification task. The regression is performed in log space (base 10). The average errors show similar performance in both our original classification and this regression case. Yet, we chose classification as the final design of our method. We make this choice because the regression network exhibits a bias towards specular materials, i.e., it performs well for specular materials, but quite poorly on diffuse materials. This becomes more evident when we look at the distribution of the estimation accuracy for shininess over the classification bins in the confusion matrix in Figure 7. The confusion matrix for the classification task (left) is symmetric at the diffuse and specular ends, whereas for the regression (right) it is more asymmetric and biased towards specular predictions. This bias is also visible on real-world data, in which case the regression network performs poorly for diffuse objects. This bias appears to result from the different losses employed in the training[4]. Please see the supplementary document for examples of this phenomenon.

Table 1. Quantitative evaluation on a test set of 4,990 synthetic images. The column 'Shininess Exponent' shows the accuracy of exponent classification, reported as percentage classified in the correct bin and the adjacent bins. The last three columns show the direct parameter estimation errors. Please note that the error on shininess is evaluated in log-space to compensate for the exponential bias.

|  | Shininess Exponent | Average Error | | |
|---|---|---|---|---|
|  | (correct bin + adjacent bins) | Diffuse Albedo | Specular Albedo | Shininess ($\log_{10}$) |
| **Our full approach** | 45.07% + 40.12% | 0.0674 | 0.2158 | 0.3073 |
| without perceptual loss (Section 5.3) | 45.15% + 40.96% | 0.1406 | 0.2368 | 0.3038 |
| without MirrorNet | 36.29% + 40.28% | 0.0759 | 0.2449 | 0.3913 |
| with exponent regression (log10) | 44.09% + 41.28% | 0.0683 | 0.2723 | 0.2974 |
| Reflectance Map Based Estimation | 13.57% + 25.29% | 0.0408 | 0.1758 | 0.7243 |

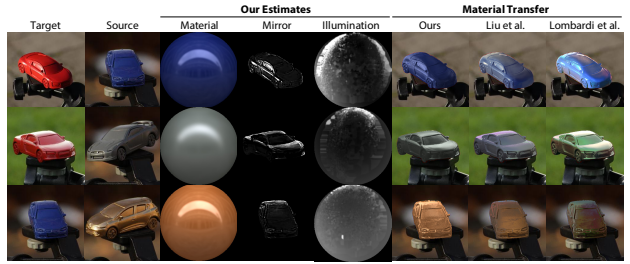

Figure 7. Confusion matrix of shininess prediction for classification (left) and regression of log-shininess (right).

---
[4]The classification task uses a binary cross-entropy loss which treats each bin as equal, whereas the regression task uses the mean absolute error, which may have greater error for larger exponent values, hence biasing.

Figure 8. Material estimation and transfer comparison. From left to right: Image of the target object, source material to copy, our estimates of material, mirror image and environment map, and the transfer results of our approach, Liu et al. [28] and Lombardi and Nishino [29]. Our method obtains better material estimates (top two rows) and illumination (third row). For fairness and comparability of the results, we use the normal map estimated by Liu et al. [28] for our environment map estimation in this case.

4. This method uses the encoder–decoder structure of Georgoulis et al. [12] to take a segmented input (from our *SegmentationNet*), and estimates a spherical reflectance map of size 256×256. This reflectance map is then fed to a second network that estimates the material parameters. Both networks are first trained independently, and then tuned by training end-to-end for a fair comparison. This approach attains slightly lower albedo errors, but performs poorly on shininess estimation. We suspect this might be due to the non-linear re-ordering required to convert an image of an object into a reflectance map, which results in the loss of high-frequency information that is essential to exponent estimation.

We train all networks on our full training data, until convergence. We also report the accuracy of our *SegmentationNet* on this test set as 99.83% (Intersection over Union). For more segmentation results, we refer to the supplementary document.

### 6.2. Comparison to the State of the Art

We compare to the state of the art in learning-based material estimation. First, we compare our material transfer results to Liu et al. [28] and Lombardi and Nishino [29] in Figure 8. The approach of Liu et al. [28] requires optimization as a post-process to obtain results of the shown quality, while our approach requires just a single forward pass of our network. Here, we also compute the environment map of the target object using the intermediate intrinsic layers regressed by our network (see Section 7). Our approach obtains more realistic material estimation and therefore better transfer results.

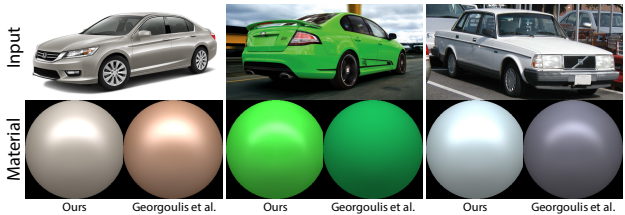

Figure 9. Comparison to the approach of Georgoulis et al. [12]. Note that their approach is specifically trained for the outdoor estimation scenario, while our approach is trained for the indoor setting. Nonetheless, our approach obtains results of similar or higher quality.

We also compare our approach to the material estimation results of Georgoulis et al. [12] in Figure 9. Please note that the shown images are taken outdoors, and their approach was trained for this specific scenario, while ours was trained for an indoor estimation setting. Nonetheless, our approach obtains results of similar or even higher quality.

### 6.3. Limitations
Our approach works well for many everyday objects, but does not handle more complex BRDFs well. Particularly difficult are scenarios that violate the assumptions that our model makes. Our approach may fail in the presence of global illumination effects such as strong shadows or inter-reflections. While most commonplace dielectric materials exhibit white specularity, some metallic objects have colored specularity, which our approach does not support. This could be addressed with more expressive BRDF and global illumination models. The quality of our material and environment map estimates depends on the quality of the input data. Modern cameras provide good white balancing, and our white illumination model hence fits well for many indoor scenarios, yet some special lighting arrangements, such as decorative lighting, require handling of color illumination. Working with low-dynamic-range images also implies dealing with camera non-linearities, which may lead to saturation artifacts, e.g., in the teddy bear in Figure 6. In our experience, the quality of surface normals derived from depth sensors is not adequate for accurate high-frequency illumination estimation. Future AR and VR devices with more advanced depth sensing capabilities may help to improve the quality of estimated environment maps.

## 7. Live Applications
Real-time material estimation from a single image or video can provide the foundation for the following exciting mixed- and augmented-reality applications:

**Single-Shot Live Material Estimation**  Our approach can estimate material parameters in a live setting, so that material properties of real-world objects can be reproduced in virtual environments from just a single image, for instance in a video game or in a VR application. For an example of such a live transfer, please see our supplementary video.

**Live Material Cloning**  When surface geometry is available, e.g., when using a depth sensor, we can extend our approach to compute an environment map alongside material parameters. This essentially converts an arbitrary real-world object into a light probe. We use the normals available from a depth sensor to map the estimated mirror image to a spherical environment map; this provides the high-frequency lighting information of the scene. We also obtain the diffuse shading image of the object and compute a low-frequency spherical harmonics lighting estimate for the scene using the available normals. The full environment map lighting is obtained by adding the two. This process is followed for single image when normals are available, for example for the target image in Figure 8. In case of a video as input, we integrate the low- and high-frequency lighting estimates of multiple time steps into the same environment map using the filtering and fusion technique described in Section 5.5. We also track the camera using the real-time volumetric VoxelHashing framework [35], so that we can integrate environment maps consistently in scene space rather than relative to the camera. We then transfer the estimate to the virtual object of our choice, and render it seamlessly into the real-world scene, as shown in Figure 10. See our supplementary video for a demonstration.

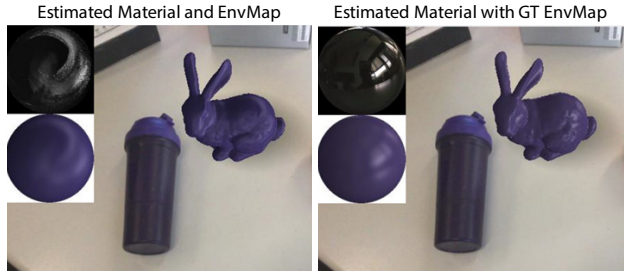

Figure 10. Cloning of real-world materials on virtual objects in an illumination-consistent fashion. **Left:** Estimated material rendered with the estimated environment map. **Right:** Estimated material rendered with the ground-truth environment map.

## 8. Conclusion
We presented the first real-time approach for estimation of diffuse and specular material appearance from a single color image. We tackle the highly complex and ill-posed inverse rendering problem using a discriminative approach based on image-to-image translation using deep encoder–decoder architectures. Our approach obtains high-quality material estimates at real-time frame rates, which enables exciting mixed-reality applications, such as illumination-consistent insertion of virtual objects and live material cloning.

We believe our approach is a first step towards real-time inverse rendering of more general materials that go beyond the commonly used Lambertian reflectance assumption and will inspire follow-up work in this exciting field.

**Acknowledgements.**  This work was supported by EPSRC grant CAMERA (EP/M023281/1), ERC Starting Grant CapReal (335545), and the Max Planck Center for Visual Computing and Communications (MPC-VCC).

# LIME: Live Intrinsic Material Estimation
# — Supplemental Document —

Abhimitra Meka [1,2]    Maxim Maximov [1,2]    Michael Zollhöfer [1,2,3]    Avishek Chatterjee [1,2]
Hans-Peter Seidel [1,2]    Christian Richardt [4]    Christian Theobalt [1,2]

[1] MPI Informatics    [2] Saarland Informatics Campus    [3] Stanford University    [4] University of Bath

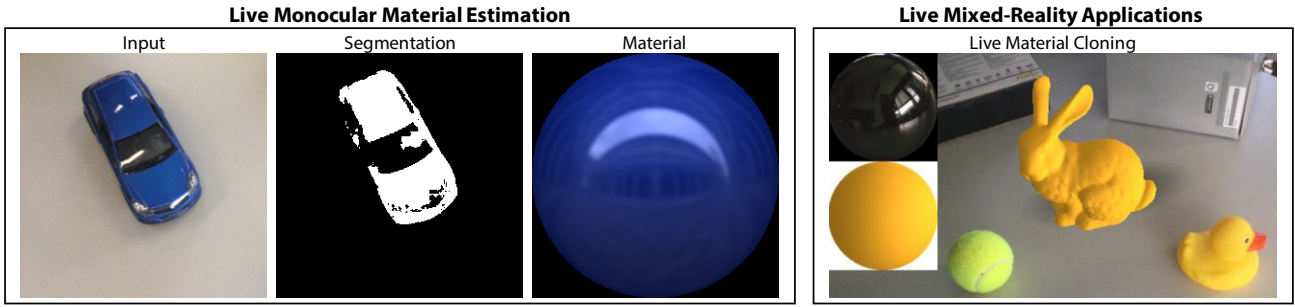

Figure 1. Our approach enables the real-time estimation of the material of general objects (left) from just a single monocular color image. This enables exciting live mixed-reality applications (right), such as for example cloning a real-world material onto a virtual object.

In this supplemental document, we show additional results, perform more evaluations, and we further justify the design choices made in our approach. More specifically, we show more results on real images (see Figure 3), more material transfer results (see Figure 4), more results on the data of Georgoulis et al. [2] (see Figure 6) and Lombardi and Nishino [4] (see Figure 7).

## 1. Shininess Classification versus Regression

As described in the main paper, our approach uses classification for recovering the specular exponent. For this classification task, we segmented the range of exponents by appearance into eight bins, which are illustrated in Figure 5. We opted for classification over regression, since regression was often overestimating the specular shininess of objects. As an example we show the material estimation results for a diffuse rubber duck in Figure 2. As is evident, the regression-based approach overestimates the specularity of the object, whereas classification is correctly able to estimate its diffuse nature.

We decided to train on synthetic training data and to sample the bins uniformly, instead of using a real-world dataset like MERL [5]. The MERL dataset consists of 100 materials. Model fits for various parametric BRDF models are available for these materials, including the Blinn–Phong model we use. The Blinn–Phong fit for these materials shows that more than 50 of the 100 material are highly specular and fall into the most specular bin in our classification system, or even higher. Sampling materials for our training data from the MERL dataset would result in our networks being heavily biased towards highly specular exponent estimation. Although a dataset such as MERL is very useful in studying how real materials span the large 4D space of BRDFs, it does not approximate well the distribution of BRDFs of everyday objects. For this reason, we chose to uniformly sample exponent values from a perceptually segmented classification system with 8 bins. For images of resolution 256×256, it is difficult to differentiate more levels of shininess, and using more bins does not provide much greater value perceptually.

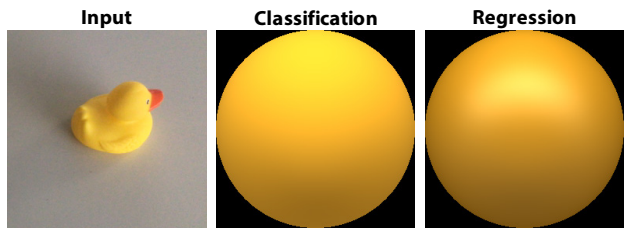

Figure 2. Regression often overestimates the shininess of objects. Therefore, we choose a classification-based approach.

## 2. Network Architecture Details

The network architecture of the five deep neural networks that make up our approach are detailed in Figures 8 to 11. SegmentationNet, SpecularNet and MirrorNet are based on the U-Net architecture [6], and AlbedoNet and ExponentNet are standard feed-forward networks for regression and classification, respectively.

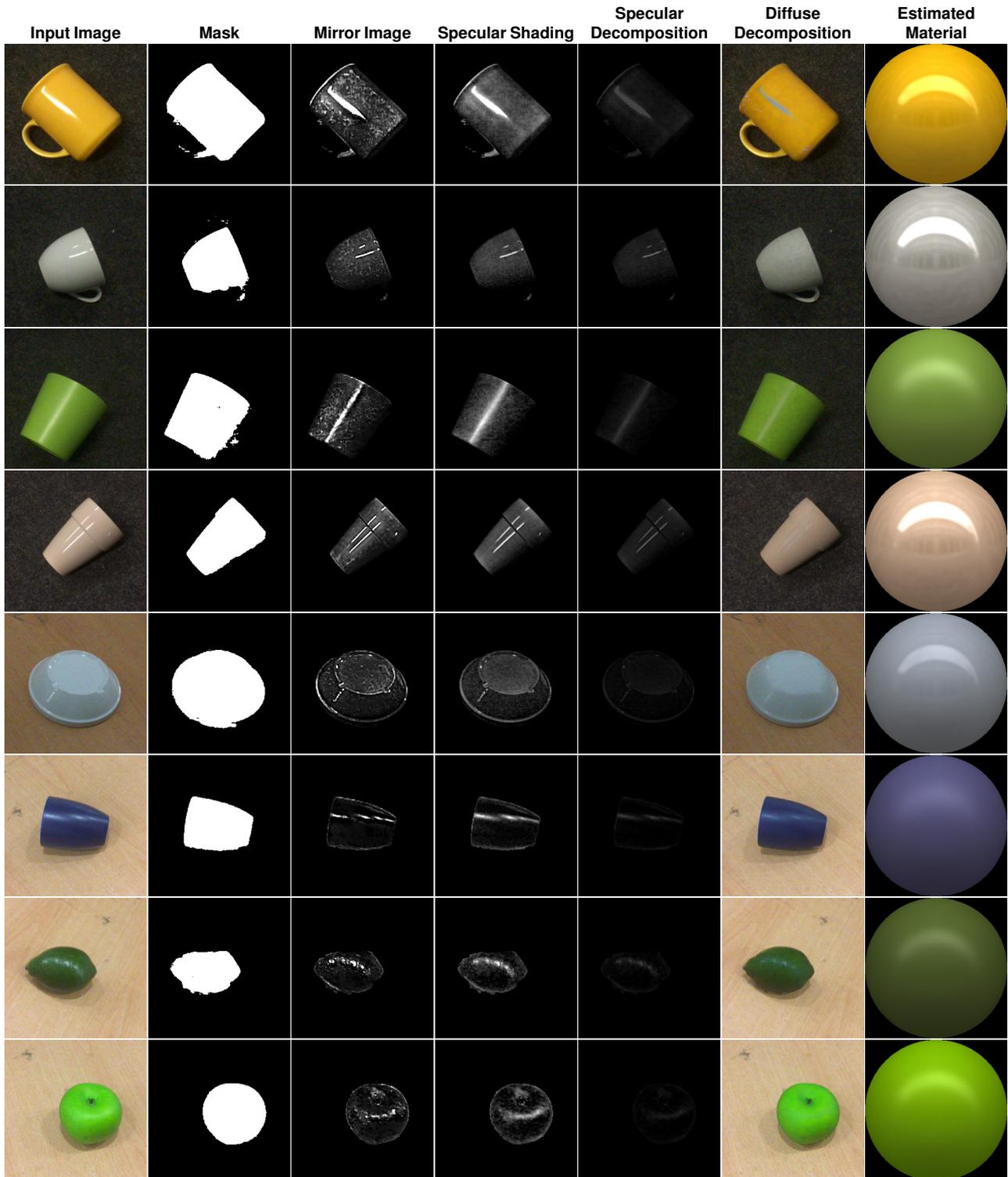

Figure 3. Real-world material estimation results based on a single color input image. Our approach produces high-quality results for a large variety of objects and materials, without manual interaction. Note the high quality of the jointly computed binary segmentation masks.

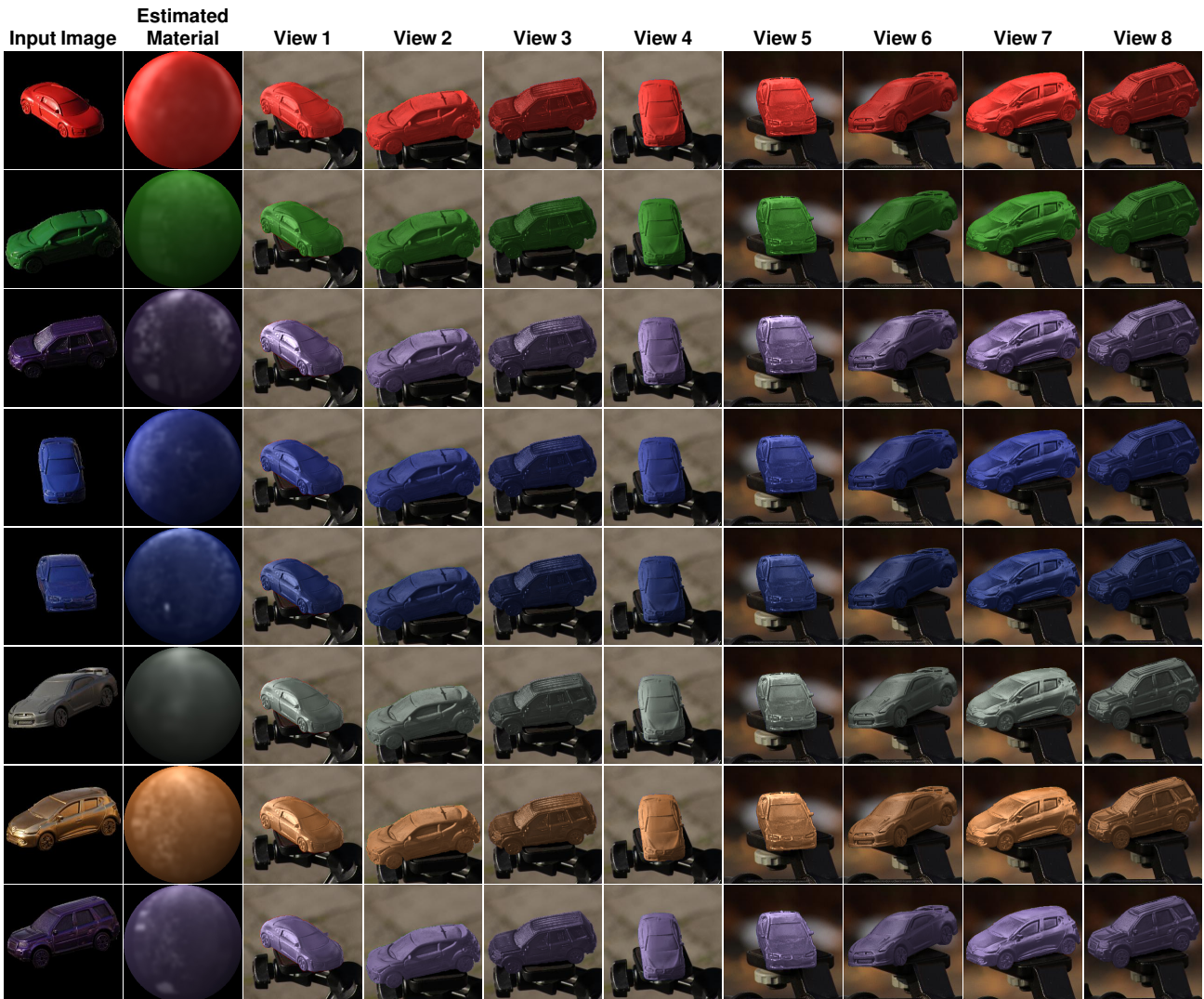

Figure 4. Material estimation and transfer. Our method obtains convincing material estimates (second column) and material transfer results (column three to ten) on the data of Liu et al. [3]. We use the normal map estimated by Liu et al. [3] as basis for our environment map estimation.

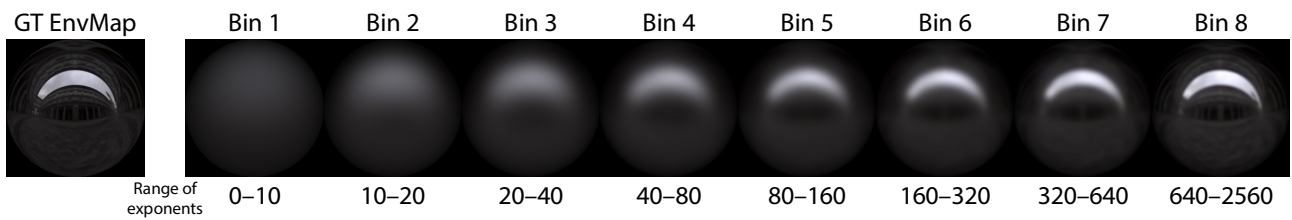

Figure 5. The 8 bins used for shininess exponent estimation in our approach, ranging from most diffuse (bin 1) to most shiny (bin 8). The visualization uses a material with a diffuse albedo of zero, a specular albedo of one, and shininess set to the mean value of each exponent bin. The materials are shown under the 'Uffizi' environment map Debevec [1].

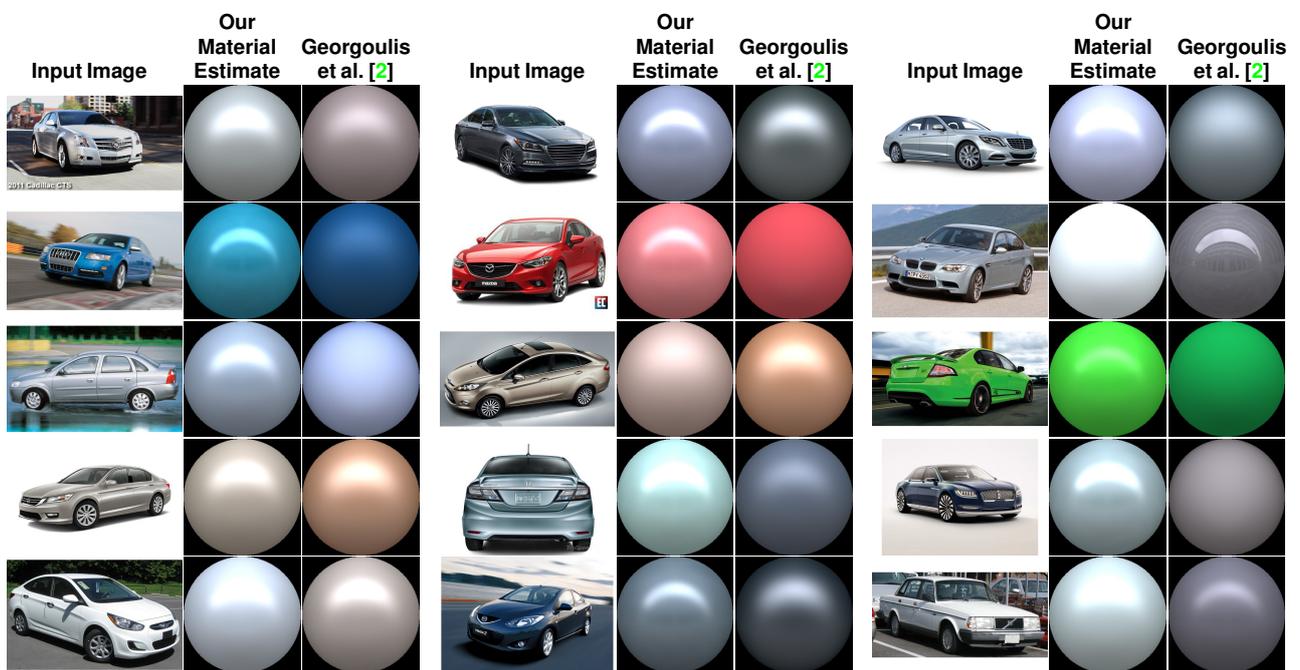

Figure 6. Comparison to the approach of Georgoulis et al. [2]. Note that their approach is specifically trained for the outdoor estimation scenario, while our approach is trained for the indoor setting. Nonetheless, our approach obtains results of similar or higher quality.

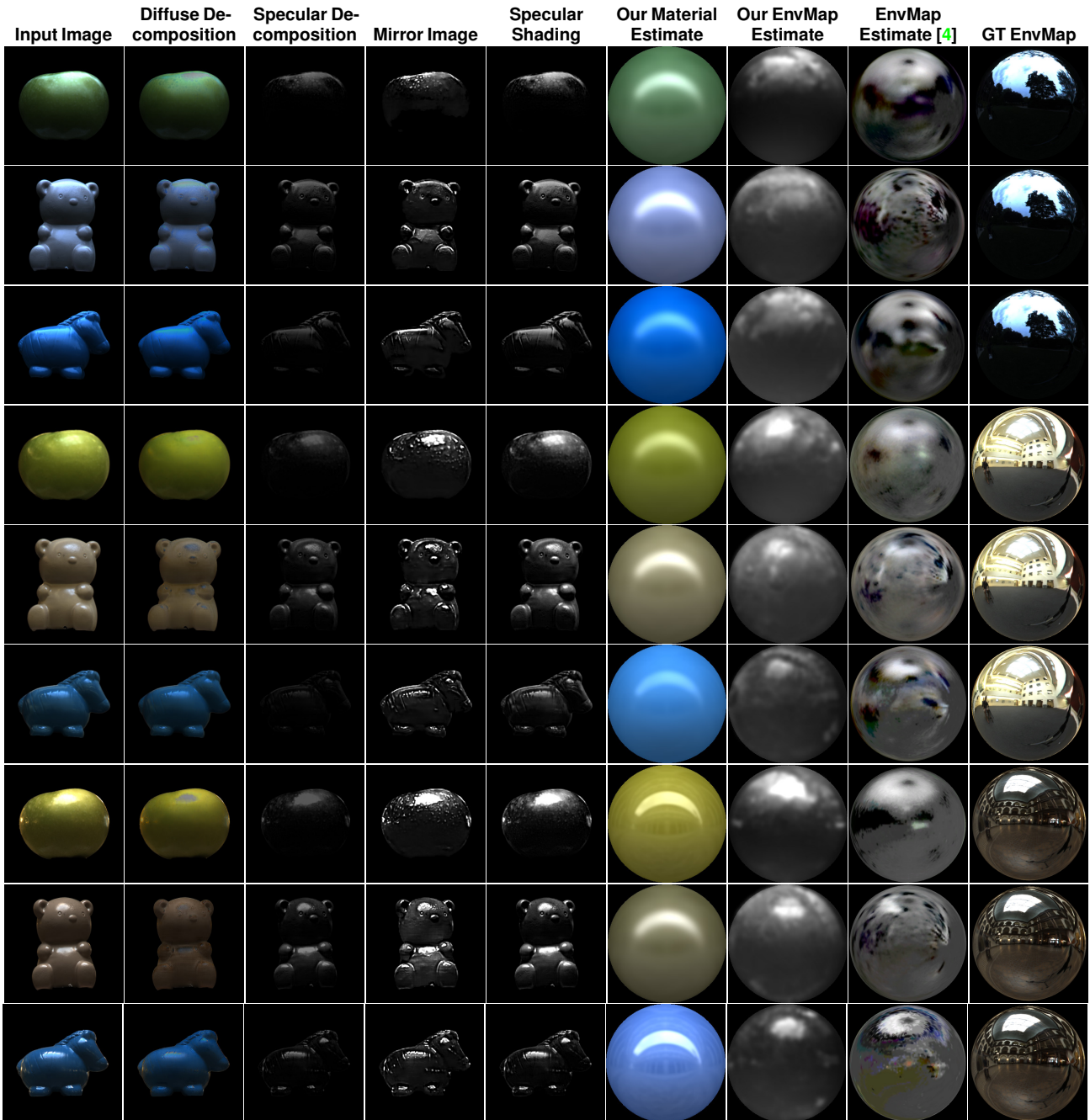

Figure 7. Our approach estimates the diffuse–specular decomposition from a single color image. The specular decomposition is obtained by multiplying specular albedo and specular shading layer, and the diffuse decomposition is obtained by subtracting the specular decomposition layer from the input image. Input images and ground-truth environment maps from Lombardi and Nishino [4]. We tone-mapped their images to process them with our approach.

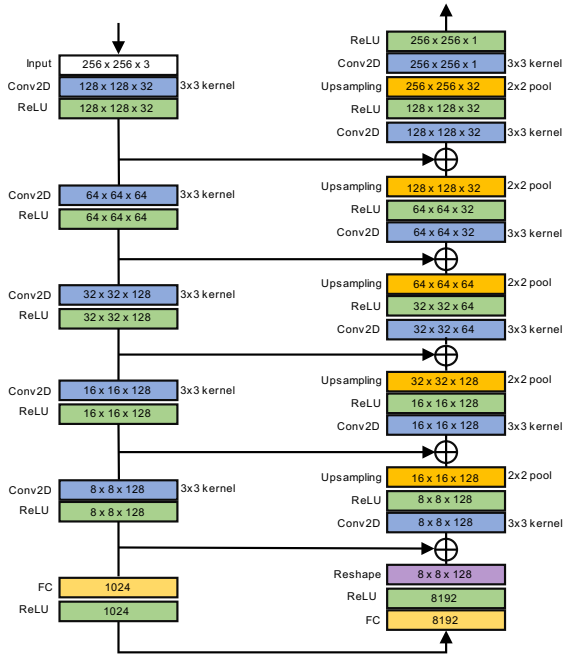

Figure 8. The architecture of our *SegmentationNet*, which learns a binary segmentation mask from a color input image. The numbers in each box denote *width×height×channels* of the layer's output, and a plus in circle represents concatenation of feature maps.

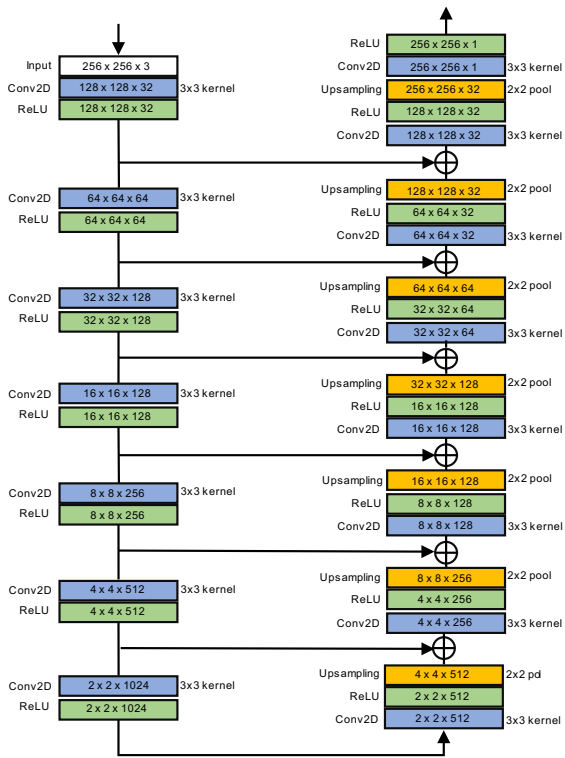

Figure 9. The architecture of our *SpecularNet*, which learns the grayscale specular decomposition from a masked color input image.

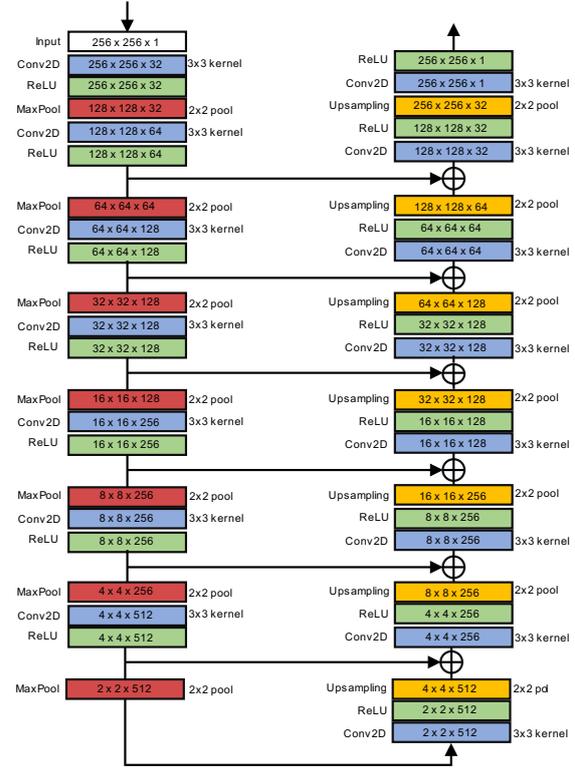

Figure 10. The architecture of our *MirrorNet*, which learns a grayscale mirror image from a grayscale specular shading image.

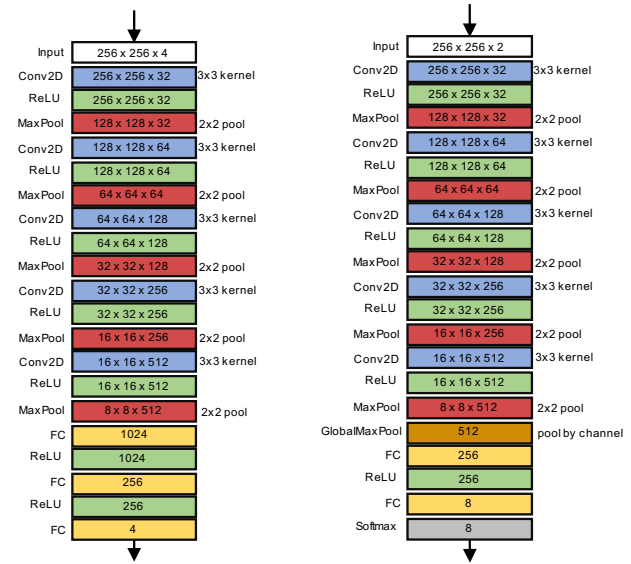

Figure 11. **Left:** The architecture of our *AlbedoNet*, which learns diffuse albedo in color (3 parameters) and grayscale specular albedo (1 parameter) from the masked color input image (3 color channels) concatenated with the grayscale specular image (1 color channel). **Right:** The architecture of our *ExponentNet*, which learns the shininess exponent using classification into 8 bins from the concatenation of the specular and mirror images (both grayscale).